\def\year{2019}\relax
\documentclass[letterpaper]{article} %DO NOT CHANGE THIS
\usepackage{aaai19}  %Required
\usepackage{times}  %Required
\usepackage{helvet}  %Required
\usepackage{courier}  %Required
\usepackage{url}  %Required
\usepackage{graphicx}  %Required

\usepackage{multirow}
\usepackage{multicol}
\usepackage{amsmath}
\usepackage{amsthm}
\usepackage{amssymb,color}
\usepackage{bm}

\title{Quantifying Uncertainties in Natural Language Processing Tasks}
\author{Yijun Xiao \and William Yang Wang\\
University of California, Santa Barbara\\
  {\tt \{yijunxiao,william\}@cs.ucsb.edu} \\
}

\newcommand{\vect}[1]{\mathbf{#1}}
\newcommand{\matr}[1]{\mathbf{#1}}

\newcommand{\vp}[0]{\vect{p}}

\newcommand{\vu}[0]{\vect{u}}

\newcommand{\vx}[0]{\vect{x}}

\newcommand{\vmu}{\boldsymbol{\mu}}
\newcommand{\vsigma}{\boldsymbol{\sigma}}
\newcommand{\vlb}{\mathcal{L}}

\newcommand{\E}{\mathbb{E}}

\newcommand{\N}{\mathcal{N}}
\newcommand{\mW}[0]{\matr{W}}

\newcommand\diag{\text{diag}}
\newcommand\softmax{\text{softmax}}

\begin{document}

\maketitle
\begin{abstract}
Reliable uncertainty quantification is a first step towards building explainable, transparent, and accountable artificial intelligent systems. Recent progress in Bayesian deep learning has made such quantification realizable. In this paper, we propose novel methods to study the benefits of characterizing model and data uncertainties for natural language processing (NLP) tasks. With empirical experiments on sentiment analysis, named entity recognition, and language modeling using convolutional and recurrent neural network models, we show that explicitly modeling uncertainties is not only necessary to measure output confidence levels, but also useful at enhancing model performances in various NLP tasks.
\end{abstract}

\section{Introduction} 
With advancement of modern machine learning algorithms and systems, they are applied in various applications that, in some scenarios, impact human wellbeing. Many of such algorithms learn black-box mappings between input and output. If the overall performance is satisfactory, these learned mappings are assumed to be correct and are used in real-life applications. It is hard to quantify how confident a certain mapping is with respect to different inputs. These deficiencies cause many AI safety and social bias issues with the most notable example being failures of auto-piloting systems. We need systems that can not only learn accurate mappings, but also quantify confidence levels or uncertainties of their predictions. With uncertainty information available, many issues mentioned above can be effectively handled.

There are many situations where uncertainties arise when applying machine learning models. First, we are uncertain about whether the structure choice and model parameters can best describe the data distribution. This is referred to as \textit{model uncertainty}, also known as epistemic uncertainty. Bayesian neural networks (BNN) \cite{buntine1991bayesian,denker1991transforming,mackay1992practical,mackay1995probable,neal2012bayesian} is one approach to quantify uncertainty associated with model parameters. BNNs represent all model weights as probability distributions over possible values instead of fixed scalars. In this setting, learned mapping of a BNN model must be robust under different samples of weights. We can easily quantify model uncertainties with BNNs by, for example, sampling weights and forward inputs through the network multiple times. Quantifying model uncertainty using a BNN learns potentially better representations and predictions due to the ensemble natural of BNNs. It is also showed in \cite{blundell2015weight} that it is beneficial for exploration in reinforcement learning (RL) problems such as contextual bandits.

Another situation where uncertainty arises is when collected data is noisy. This is often the case when we rely on observations and measurements to obtain the data. Even when the observations and measurements are precise, noises might exist within the data generation process. Such uncertainties are referred to as \textit{data uncertainties} in this paper and is also called aleatoric uncertainty \cite{der2009aleatory}. Depending on whether the uncertainty is input independent, data uncertainty is further divided into \textit{homoscedastic uncertainty} and \textit{heteroscedastic uncertainty}. Homoscedastic uncertainty is the same across the input space which can be caused by systematic observation noise. Heteroscedastic uncertainty, on the contrary, is dependent on the input. For example, when predicting the sentiment of a Yelp review, single-word review ``good'' is possible to have 3, 4 or 5-star ratings while a lengthened review with strong positive emotion phrases is definitely a 5-star rating. In the rest of the paper, we also refer to heteroscedastic uncertainty as input-dependent data uncertainty.

Recently, there are increasing number of studies investigating the effects of quantifying uncertainties in different applications \cite{kendall2015bayesian,gal2016theoretically,kendall2017uncertainties,zhu2017deep}. In this paper, we focus on exploring the benefits of quantifying both model and data uncertainties in the context of various natural language processing (NLP) tasks. Specifically, we study the effects of quantifying model uncertainty and input-dependent data uncertainty in sentiment analysis, named entity recognition, and language modeling tasks. We show that there is a potential performance increase when including both uncertainties in the model. We also analyze the characteristics of the quantified uncertainties.

The main contributions of this work are:
\begin{enumerate}
\item We mathematically define model and data uncertainties via the law of total variance;
\item Our empirical experiments show that by accounting for model and data uncertainties, we observe significant improvements in three important NLP tasks;
\item We show that our model outputs higher data uncertainties for more difficult predictions in sentiment analysis and named entity recognition tasks.
\end{enumerate}
%\william{When you write the intro, please start with a big picture: e.g., the ultimate goal is that we care about many different attributes (e.g., explainability, transparency, accuracy, accountability, bias etc) about ML models. Here we'd like to explore if modeling uncertainty explicitly could improve the accuracy of NLP problems.}

\section{Related Work}

\subsection{Bayesian Neural Networks}
Modern neural networks are parameterized by a set of model weights $\mW$. In the supervised setting, for a dataset $D=\{(\vx_1,y_i)\}_{i=1}^N$, a point estimate for $\mW$ is obtained by maximizing certain objective function. Bayesian neural networks \cite{buntine1991bayesian,denker1991transforming,mackay1992practical,mackay1995probable,neal2012bayesian} introduce model uncertainties by putting a prior on the network parameters $p(\mW)$. Bayesian inference is adopted in training aiming to find the posterior distribution of the parameters $p(\mW|D)$ instead of a point estimate. This posterior distribution describes possible values for the model weights given the dataset. Predictive function $f^\mW(\vx)$ is used to predict the corresponding $y$ value. Given the posterior distribution for $\mW$, the function is marginalized over $\mW$ to obtain the expected prediction.

Exact inference for BNNs is rarely available given the complex nonlinear structures and high dimension of model parameters $\mW$ of modern neural networks. Various approximate inference methods are proposed \cite{graves2011practical,hernandez2015probabilistic,blundell2015weight,gal2016dropout}. In particular, Monte Carlo dropout (MC dropout) \cite{gal2016dropout} requires minimum modification to the original model. Dropouts are applied between nonlinearity layers in the network and are activated at test time which is different from a regular dropout. They showed that this process is equivalent to variational Bayesian approximation where the approximating distribution is a mixture of a zero mean Gaussian and a Gaussian with small variances. When sampling dropout masks, model outputs can be seen as samples from the posterior predictive function $f^{\widehat{\mW}}(\vx)$ where $\widehat{\mW}\sim p(\mW|D)$. As a result, model uncertainty can be approximately evaluated by finding the variance of the model outputs from multiple forward passes.

\subsection{Uncertainty Quantification}
Model uncertainty can be quantified using BNNs which captures uncertainty about model parameters. Data uncertainty describes noises within the data distribution. When such noises are homogeneous across the input space, it can be modeled as a parameter. In the cases where such noises are input-dependent, i.e. observation noise varies with input $\vx$, heteroscedastic models \cite{nix1994estimating,le2005heteroscedastic} are more suitable. 

Recently, quantifications of model and data uncertainties are gaining researchers' attentions. Probabilistic pixel-wise semantic segmentation has been studied in \cite{kendall2015bayesian}; Gal and Ghahramani \shortcite{gal2016theoretically} studied model uncertainty in recurrent neural networks in the context of language modeling and sentiment analysis; Kendall and Gal \shortcite{kendall2017uncertainties} researched both model and data uncertainties in various vision tasks and achieved higher performances; Zhu and Laptev \shortcite{zhu2017deep} used similar approaches to perform time series prediction and anomaly detection with Uber trip data. This study focuses on the benefits of quantifying model and data uncertainties with popular neural network structures on various NLP tasks.

\section{Methods}
First of all, we start with the law of total variance. Given a input variable $x$ and its corresponding output variable $y$, the variance in $y$ can be decomposed as:
\begin{align}
\label{equ:total_var}
\text{Var}(y)=\text{Var}\left(\E[y|x]\right)+\E\left[\text{Var}(y|x)\right]
\end{align}

We mathematically define model uncertainty and data uncertainty as:
\begin{align}
\label{equ:def_um}
U_m(y|x)&=\text{Var}\left(\E[y|x]\right) \\
\label{equ:def_ui}
U_d(y|x)&=\E\left[\text{Var}(y|x)\right]
\end{align}
where $U_m$ and $U_d$ are model and data uncertainties respectively. We can see that both uncertainties partially explain the variance in the observation. In particular, model uncertainty explains the part related to the mapping process $\E[y|x]$ and data uncertainty describes the variance inherent to the conditional distribution $\text{Var}(y|x)$. By quantifying both uncertainties, we essentially are trying to explain different parts of the observation noise in $y$.

In the following sections, we introduce the methods employed in this study to quantify uncertainties.

\begin{figure*}\centering    
	\includegraphics[width=6.4in]{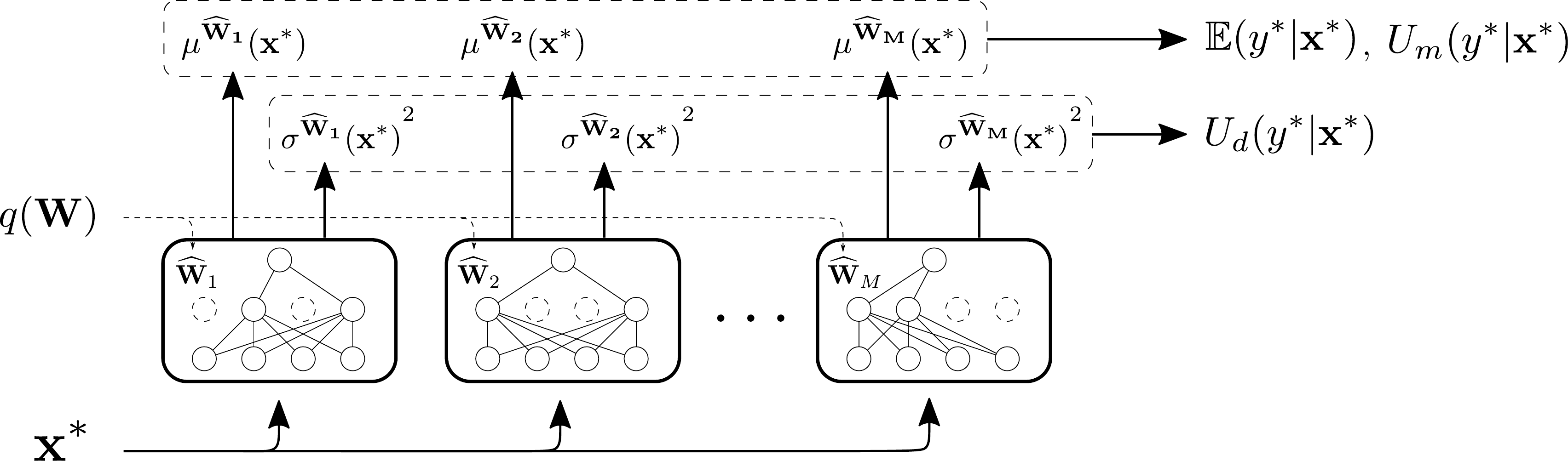}
	\caption{\label{fig:uncertainty_eval}Illustration of the evaluation process of predicted output and both model uncertainty and data uncertainty. $\E(y^*|\vx^*)$ denotes the expected value of model prediction; $U_m(y^*)$ is the model uncertainty with respect to the output; $U_d(y^*)$ is the input-dependent data uncertainty. Dotted arrows represent sampling processes.}
\end{figure*}

\subsection{Model Uncertainty}
Recall that Bayesian neural networks aim to find the posterior distribution of $\mW$ given the dataset $D=\{(\vx_1,y_i)\}_{i=1}^N$. We also specify the data generating process in the regression case as:
\begin{align}
y|\mW \sim \N(f^\mW(\vx), \sigma^2)
\end{align}

With the posterior distribution $p(\mW|D)$, given a new input vector $\vx^*$, the prediction is obtained by marginalizing over the posterior:

\begin{align}
\label{equ:prediction}
p(y^*|\vx^*,D)=\int_\mW{p\left(y^*|f^\mW(\vx^*)\right)p(\mW|D)d\mW}
\end{align}

As exact inference is intractable in this case, we can use variational inference approach to find an approximation $q_\theta(\mW)$ to the true posterior $p(\mW|D)$ parameterized by a different set of weights $\theta$ where the Kullback-Leibler (KL) divergence of the two distributions is minimized.

There are several variational inference methods proposed for Bayesian neural networks \cite{hernandez2015probabilistic,blundell2015weight,gal2016dropout}. In particular, dropout variational inference method \cite{gal2016dropout}, when applied to models with dropout layers, requires no retraining and can be applied with minimum changes. The only requirement is dropouts have to be added between nonlinear layers. At test time, dropouts are activated to allow sampling from the approximate posterior. We use MC dropout in this study to evaluate model uncertainty.

%\william{Why? Some motivations? What if reviewers claim that this is just regularization?} 

At test time, we have the optimized approximated posterior $q(\mW)$. Prediction distribution can be approximated by switching $p(\mW|D)$ to $q(\mW)$ in Equation \ref{equ:prediction} and perform Monte Carlo integration as follows:
\begin{align}
\E(y^*|\vx^*)\approx\frac{1}{M}\sum_{j=1}^Mf^{\widehat{\mW}_j}(\vx^*)
\end{align}
Predictive variance can also be approximated as:
\begin{align}
\label{equ:var1}
\text{Var}\left(y^*\right)\approx \frac{1}{M}\sum_{j=1}^Mf^{\widehat{\mW}_j}(\vx^*)^2-\E(y^*|\vx^*)^2 +\sigma^2
\end{align}
where $\widehat{\mW}_j$ is sampled from $q(\mW)$.

Note here $\sigma^2$ is the inherent noise associated with the inputs which is homogeneous across the input space. This is often considered by adding a weight decay term in the loss function. We will discuss the modeling of input-dependent data uncertainty in the next section. The rest part of the variance arises because of the uncertainty about the model parameters $\mW$. We use this to quantify model uncertainty in the study, i.e.:
\begin{align}
U_m(y^*|\vx^*)=\frac{1}{M}\sum_{j=1}^Mf^{\widehat{\mW}_j}(\vx^*)^2-\E(y^*|\vx^*)^2
\end{align}
%\william{What is the relationship between input uncertainty and model (parameter) uncertainty? Not sure if this is the best place to discuss this, but we can also discuss this in the later sections.}

\begin{table*}[t!]
\begin{center}
\begin{tabular}{lccccc}
\hline
{\bf Corpus} & \bf Size & \bf Average Tokens & $|V|$ & \bf Classes & \bf Class Distribution\\ 
\hline
Yelp 2013 & 335,018   & 151.6 & 211,245 & 5 & .09/.09/.14/.33/.36 \\
Yelp 2014 & 1,125,457 & 156.9 & 476,191 & 5 & .10/.09/.15/.30/.36 \\
Yelp 2015 & 1,569,264 & 151.9 & 612,636 & 5 & .10/.09/.14/.30/.37\\
IMDB & 348,415 & 325.6 & 115,831 & 10 & .07/.04/.05/.05/.08/.11/.15/.17/.12/.18\\ 

\hline
\end{tabular}
\end{center}
\caption{\label{tab:sent_datasets} Summaries of Yelp 2013/2014/2015 and IMDB datasets. $|V|$ represents the vocabulary size.}
\end{table*}

\subsection{Data Uncertainty}
\label{sec:input_uncertainty}
Data uncertainty can be either modeled homogeneous across input space or input-dependent. We take the second option and make the assumption that data uncertainty is dependent on the input. To achieve this, we need to have a model that not only predicts the output values, but also estimates the output variances given some input. In other words, the model needs to give an estimation of $\text{Var}(y|x)$ mentioned in Equation \ref{equ:def_ui}.
%\william{I sort of understand your motivation, but can we explain better why measuring output variances is needed for understanding input variances? Is this a direct or indirect measure? Why?} 

Denote $\mu(\vx)$ and $\sigma(\vx)$ as functions parameterized by $\mW$ that calculate output mean and standard deviation for input $\vx$ (in practice, logarithm of the variance is calculated for an improvement on stability). We make the following assumption on the data generating process:
\begin{align}
y\sim\N\left(\mu(\vx), \sigma(\vx)^2\right)
\end{align}

Given the setting and the assumption, the negative data log likelihood can be written as follows:

\begin{align}
\label{equ:rgs_loss}
\vlb_{\text{rgs}}(\mW)= & -\frac{1}{N}\sum_{i=1}^N\log{p(y_i|\mu(\vx_i),\sigma(\vx_i))} \nonumber \\
= & \frac{1}{N} \sum_{i=1}^N \left(\frac{1}{2}\left|\frac{y_i-\mu(\vx_i)}{\sigma(\vx_i)}\right|^2 + \right.\nonumber \\
& \left.\frac{1}{2}\log{\sigma(\vx_i)^2} + \frac{1}{2}\log{2\pi}\right)
\end{align}

Comparing Equation \ref{equ:rgs_loss} to a standard mean squared loss used in regression, we can see that the model encourages higher variances estimated for inputs where the predicted mean $\mu(\vx_i)$ is more deviated from the true observation $y_i$. On the other hand, a regularization term on the $\sigma(\vx_i)$ prevents the model from estimating meaninglessly high variances for all inputs. Equation \ref{equ:rgs_loss} is referred to as learned loss attenuation in \cite{kendall2017uncertainties}.

While Equation \ref{equ:rgs_loss} works desirably for regression, it is based on the assumption that $y\sim\N(\mu(\vx), \sigma(\vx)^2)$. This assumption clearly does not hold in the classification context. We can however adapt the same formulation in the logit space. In detail, define $\vmu(\vx)$ and $\vsigma(\vx)$ as functions that maps input $\vx$ to the logit space. Logit vector is sampled and thereafter transformed into probabilities using softmax operation. This process can be described as:

\begin{align}
\vu &\sim \N\left(\vmu(\vx), \diag(\vsigma(\vx)^2)\right) \\
\vp &= \softmax(\vu) \\
y &\sim \text{Categorical}(\vp) \label{equ:y_i}
\end{align}
where $\diag()$ function takes a vector and output a diagonal matrix by putting the elements on the main diagonal. 
Note here in Equation \ref{equ:y_i}, $y$ is a single label. This formulation can be easily extended to multi-way Categorical labels.

During training, we seek to maximize the expected data likelihood. Here we approximate the expected distribution for $\vp$ using Monte Carlo approximation as follows:

\begin{align}
\vu^{(k)}& \sim \N\left(\vmu(\vx), \diag(\vsigma(\vx)^2)\right) \\
\E[\vp]& \approx\frac{1}{K}\sum_{k=1}^K\softmax(\vu^{(k)})
\end{align}

The negative log-likelihood for the dataset can be written as:

\begin{align}
\label{equ:clf_loss}
\vlb_{\text{clf}}(\mW)= &\frac{1}{N}\sum_{i=1}^N\log\sum_{k=1}^K\exp\left(u_{i,y_i}^{(k)}-\log\sum_c\exp{u_{i,c}^{(k)}}\right) \nonumber\\
 & - \log{K}
\end{align}
where $u_{i,c}$ is the $c$-th element in $\vu_i$.

After the model is optimized, we use $\sigma(\vx^*)^2$ to estimate the data uncertainty given input $\vx^*$ in the regression case:
\begin{align}
U_d(y^*|\vx^*)=\sigma(\vx^*)^2
\end{align}

For classification, we use the average variance of the logits as a surrogate to quantify the data uncertainty. This does not directly measures data uncertainty in the output space but can reflect to a certain extent the variance caused by the input.

\subsection{Combining Both Uncertainties}
To simultaneously quantify both uncertainties, we can simply use Equation \ref{equ:rgs_loss},\ref{equ:clf_loss} in the training stage and adopt MC dropout during evaluation as described in the model uncertainty section.

Take the regression setting as an example, prediction can be approximated as:
\begin{align}
\E(y^*|\vx^*)\approx\frac{1}{M}\sum_{j=1}^M\mu^{\widehat{\mW}_j}(\vx^*)
\end{align}
Model uncertainty can be measured with:
\begin{align}
U_m(y^*|\vx^*)=\frac{1}{M}\sum_{j=1}^M\mu^{\widehat{\mW}_j}(\vx^*)^2-\E(y^*|\vx^*)^2
\end{align}
and data uncertainty is quantified with:
\begin{align}
\label{equ:ui}
U_d(y^*|\vx^*)=\frac{1}{M}\sum_{j=1}^M\sigma^{\widehat{\mW}_j}(\vx^*)^2
\end{align}
where again $\widehat{\mW}_j$ is sampled from $q(\mW)$. Figure \ref{fig:uncertainty_eval} is an illustration of the evaluation process of predictive value and different uncertainty measures.
%\william{Overall, my comment for the method section is that we need to provide more justifications and motivations of the chosen method. These would help reviewers understand your approach.}

\section{Experiments and Results}
We conduct experiments on three different NLP tasks: sentiment analysis, named entity recognition, and language modeling. In the following sections, we will introduce the datasets, experiment setups, evaluation metrics for each task, and experimental results.

\begin{table*}[t!]
\begin{center}
\begin{tabular}{lcccc}
\hline
{\bf Model} & {\bf Yelp 2013} & {\bf Yelp 2014} & {\bf Yelp 2015} & {\bf IMDB}\\  \hline
% \textsc{clf acc} \\
% Yelp 2013 & 61.9 & 62.3 & 61.5 & \bf 62.3 \\
% Yelp 2014 & 63.6 & 63.8 & 63.3 & \bf 63.8 \\
% Yelp 2015 & 64.2 & 64.4 & 64.2 & \bf 64.6 \\
% IMDB      & 36.7 & 37.0 & 36.5 & \bf 37.1 \\ 
% \hline
(\textsc{rgs mse}) \\
Baseline & 0.71 & 0.72 & 0.72 & 3.62 \\
Baseline + \textsc{mu} & 0.57 & 0.55 & 0.55 & 3.20\\
Baseline + \textsc{du} & 0.84 & 0.75 & 0.73 & 3.74 \\
Baseline + both        & \bf 0.57 & \bf 0.54 & \bf 0.53 & \bf 3.13 \\ \hline
Relative Improvement (\%) & 19.7 & 25.0 & 26.4 & 13.5\\
\hline
\end{tabular}
\end{center}
\caption{\label{tab:sent_results}Test set mean squared error of CNN regressors trained on four sentiment analysis datasets. \textsc{rgs mse} represents regression MSE. Baseline is the baseline CNN model \cite{kim2014convolutional}; \textsc{mu} and \textsc{du} denote model uncertainty and data uncertainty respectively. Classification results have a similar pattern but the improvements are less obvious.
%\william{I'm worried that reviewers won't read the second half of the table or the text carefully, and then they say the improvements are marginal. One strategy: we only show the text regression experiment results and relative improvements in details in this table, but say that we don't observe much improvements for the classification setting in the text, because the search space (related to uncertainty) for regression is much larger than the limited categories in classification.}
}
\end{table*}

\subsection{Sentiment Analysis}
Conventionally, sentiment analysis is done with classification. In this study, to explore the effect of quantifying uncertainties, we consider both regression and classification settings for sentiment analysis. In the regression setting, we treat the class labels as numerical values and aim to predict the real value score given a review document. We introduce the datasets and setups in both settings in this section.

\paragraph{Datasets} We use four large scale datasets containing document reviews as in \cite{tang2015document}. Specifically, we use IMDB movie review data \cite{diao2014jointly} and Yelp restaurant review datasets
from Yelp Dataset Challenge in 2013, 2014 and 2015. Summaries of the four datasets are given in Table \ref{tab:sent_datasets}. Data splits are the same as in \cite{tang2015document,diao2014jointly}.

\paragraph{Experiment Setup} We implement convolutional neural network (CNN) baselines in both regression and classification settings. CNN model structure follows \cite{kim2014convolutional}. We use a maximum vocabulary size of 20,000; embedding size is set to 300; three different kernel sizes are used in all models and they are chosen from [(1,2,3), (2,3,4), (3,4,5)]; number of feature maps for each kernel is 100; dropout \cite{srivastava2014dropout} is applied between layers and dropout rate is 0.5. To evaluate model uncertainty and input uncertainty, 10 samples are drawn from the approximated posterior to estimate the output mean and variance.

Adam \cite{kingma2014adam} is adopted in all experiments with learning rate chosen from [3e-4, 1e-3, 3e-3] and weight decay from [3e-5, 1e-4, 3e-4]. Batch size is set to 32 and training runs for 48 epochs with 2,000 iterations per epoch for Yelp 2013 and IMDB, and 5,000 iterations per epoch for Yelp 2014 and 2015. Model with best performance on the validation set is chosen to be evaluated on the test set.

\paragraph{Evaluation}
We use accuracy in the classification setting and mean squared error (MSE) in the regression setting to evaluate model performances. Accuracy is a standard metric to measure classification performance. MSE measures the average deviation of the predicted scores from the true ratings and is defined as:
\begin{align}
\text{MSE}=\frac{\sum_{i=1}^N(\text{gold}_i-\text{predicted}_i)^2}{N}
\end{align}

\paragraph{Results} Experiment results are shown in Table \ref{tab:sent_results}. We can see that BNN models (i.e. model w/ \textsc{mu} and w/ both) outperform non-Bayesian models. Quantifying both model and data uncertainties boosts performances by 13.5\%-26.4\% in the regression setting. Most of the performance gain is from quantifying model uncertainty. Modeling input-dependent uncertainty alone marginally hurts prediction performances. The performances for classification increase marginally with added uncertainty measures. We conjecture that this might be due to the limited output space in the classification setting. 

\begin{figure}\centering    
	\includegraphics[width=3.2in]{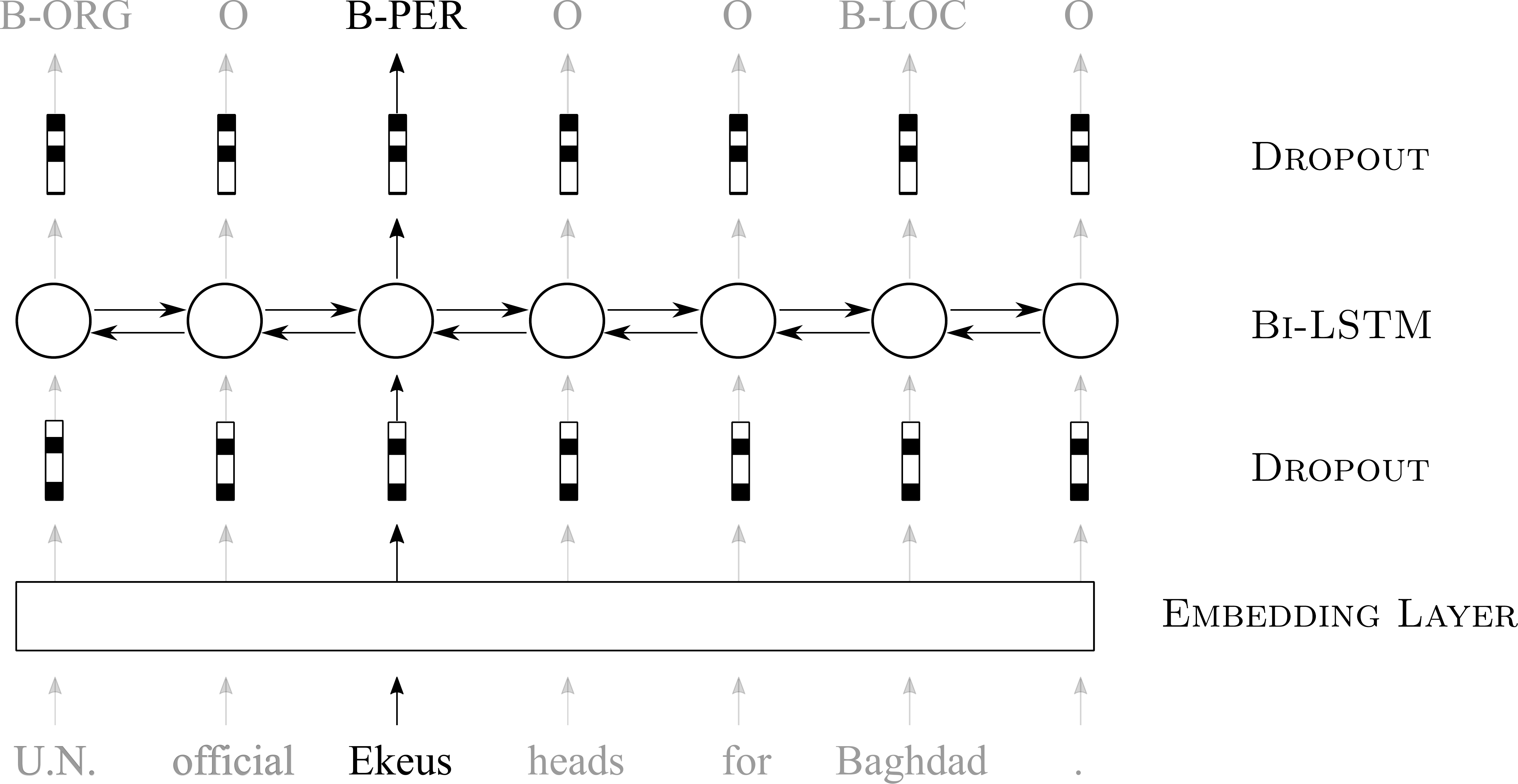}
	\caption{An illustration of the bidirectional LSTM model used for named entity recognition. Two dropout layers independently sample their masks while masks are the same across time steps.\label{fig:bilstm}}
\end{figure}

\subsection{Named Entity Recognition}
We conduct experiments on named entity recognition (NER) task which essentially is a sequence tagging problem. We adopt a bidirectional long-short term memory (LSTM) \cite{hochreiter1997long} neural network as the baseline model and measure the effects of quantifying model and input-dependent uncertainties on the test performances.

\paragraph{Datasets} For the NER experiments, we use the CoNLL 2003 dataset \cite{tjong2003introduction}. This corpus consists of news articles from the Reuters RCV1 corpus annotated with four types of named entities: location, organization, person, and miscellaneous. The annotation scheme is IOB (which stands for inside, outside, begin, indicating the position of the token in an entity). The original dataset includes annotations for part of speech (POS) tags and chunking results, we do not include these features in the training and use only the text information to train the NER model.

\begin{table}[t!]
\begin{center}
\begin{tabular}{lc}
\hline
{\bf Model} & {\bf CoNLL 2003}\\  \hline
(\textsc{f1 score}) \\
Baseline & 77.5 \\
Baseline + \textsc{mu} & 76.5\\
Baseline + \textsc{du} & \bf 79.6 \\
Baseline + both        & 78.5 \\ \hline
Relative Improvement (\%) & 2.7 \\
\hline
\end{tabular}
\end{center}
\caption{\label{tab:ner_results}Test set F1 scores (\%) of bidirectional LSTM taggers trained on CoNLL 2003 dataset. Baseline is the baseline bidirectional LSTM model; \textsc{mu} and \textsc{du} denote model uncertainty and data uncertainty respectively. Modeling data uncertainty boosts performances}
\end{table}

\paragraph{Experiment Setup} Our baseline model is a bidirectional LSTM with dropout applied after the embedding layer and before the output layer. We apply dropout with the same mask for all time steps following \cite{gal2016theoretically}. An illustration of the model is shown in Figure \ref{fig:bilstm}. Note that the dropout mask is the same across time steps. Different examples in the same mini-batch have different dropout masks.

Word embedding size is 200 and hidden size in each direction is 200; dropout probability is fixed at 0.5; other hyper-parameters related to quantifying uncertainties are the same with previous experiment setups.

For training, we use Adam optimizer \cite{kingma2014adam}. Learn rate is selected from [3e-4, 1e-3, 3e-4] and weight decay is chosen from [0, 1e-5, 1e-4]. Training runs for 100 epochs with each epoch consisting of 2,000 randomly sampled mini-batches. Batch size is 32.

\paragraph{Evaluation}
The performances of the taggers are measured with F1 score:
\begin{align}
\text{F1}=\frac{2\cdot\text{precision}\cdot\text{recall}}{\text{precision}+\text{recall}}
\end{align}
where precision is the percentage of entities tagged by the model that are correct; recall is the percentage of entities in the gold annotation that are tagged by the model. A named entity is correct only if it is an exact match of the corresponding entity in the data.

\paragraph{Results}
Test set performances of the models trained with and without uncertainties are listed in Table \ref{tab:ner_results}. We observe that much different from the sentiment analysis case, models that quantify data uncertainty improves performances by 2.7\% in F1 score. Quantifying model uncertainty, on the other hand, under-performs by approximately 1\% absolute F1 score. One possible explanation for worse results with model uncertainty is due to the use of MC dropout and chunk based evaluation. More specifically, predicted tag at each time step is taken to be the argmax of the average tag probability across multiple passes with the same inputs. This operation might break some temporal dynamics captured with a single pass of the inputs.

\begin{table}[t!]
\begin{center}
\begin{tabular}{lc}
\hline
{\bf Model} & {\bf PTB}\\  \hline
(\textsc{ppl}) \\
Baseline & 82.7 \\
Baseline + \textsc{mu} & 81.3\\
Baseline + \textsc{du} & 80.5 \\
Baseline + both        & \bf 79.2 \\ \hline
Relative Improvement (\%) & 4.2 \\
\hline
\end{tabular}
\end{center}
\caption{\label{tab:lm_results}Test set perplexities of LSTM language models trained on PTB dataset. \textsc{ppl} represents perplexity. Baseline is the baseline medium two-layer LSTM model in \cite{zaremba2014recurrent}; \textsc{mu} and \textsc{du} denote model uncertainty and data uncertainty respectively.}
\end{table}

\subsection{Language Modeling}
We introduce the experiments conducted on the language modeling task. 

\paragraph{Datasets} We use the standard Penn Treebank (PTB), a standard benchmark in the field. The dataset contains 887,521 tokens (words) in total.

\paragraph{Experiment Setting} We follow the medium model setting in \cite{zaremba2014recurrent}. The model is a two-layer LSTM with hidden size 650. Dropout rate is fixed at 0.5. Dropout is applied after the embedding layer, before the output layer, and between two LSTM layers. Similar to the NER setting, dropout mask is the same across time steps. Unlike \cite{gal2016theoretically}, we do not apply dropout between time steps. Weight tying is also not applied in our experiments. Number of samples for MC dropout is set to 50.

\paragraph{Evaluation} We use the standard perplexity to evaluate the trained language models.

\paragraph{Results} The results are shown in Table \ref{tab:lm_results}. We can observe performance improvements when quantifying either model uncertainty or data uncertainty. We observe less performance improvements  compared to \cite{gal2016theoretically} possibly due to the fact that we use simpler dropout formulation that only applies dropout between layers.

\begin{table}[t!]
\begin{center}
\begin{tabular}{p{3.in}}
\hline
{\bf High \textsc{du}}  \\  \hline
should game automatic doors ! \\
\hline
i 've bought tires from discount tire for years at different locations and have had a \textit{good} experience , but this location was different . i went in to get some new tires with my fiancé . john the sales guy pushed a certain brand , specifically because they were running a rebate special . tires are tires , especially on a prius (the rest 134 tokens not shown here due to space) \\
\hline
{\bf Low \textsc{du}} \\ \hline
\textit{great} sports bar ! brian always goes out of his way to make sure we are \textit{good} to go ! \textit{great} people , \textit{great} food , \textit{great} music ! \textit{great} bartenders and even \textit{great} bouncers ! always accommodating ! all the \textit{best} \_unk ! \\ \hline
\textit{great} \_unk burger ! \textit{amazing} service ! \textit{brilliant} interior ! the burger was \textit{delicious} but it was a little big . it 's a \textit{great} restaurant \textit{good} for any occasion .\\ \hline
\end{tabular}
\end{center}
\caption{\label{tab:sent_iu}Examples of inputs in Yelp 2013 dataset with high and low data uncertainties. They are taken from the top and bottom 10 examples with respect to measured data uncertainty. High \textsc{du} is around 0.80 and low is around 0.52. Italic tokens are highly indicative tokens for higher ratings.}
\end{table}

\begin{figure}\centering    
	\includegraphics[width=3.2in]{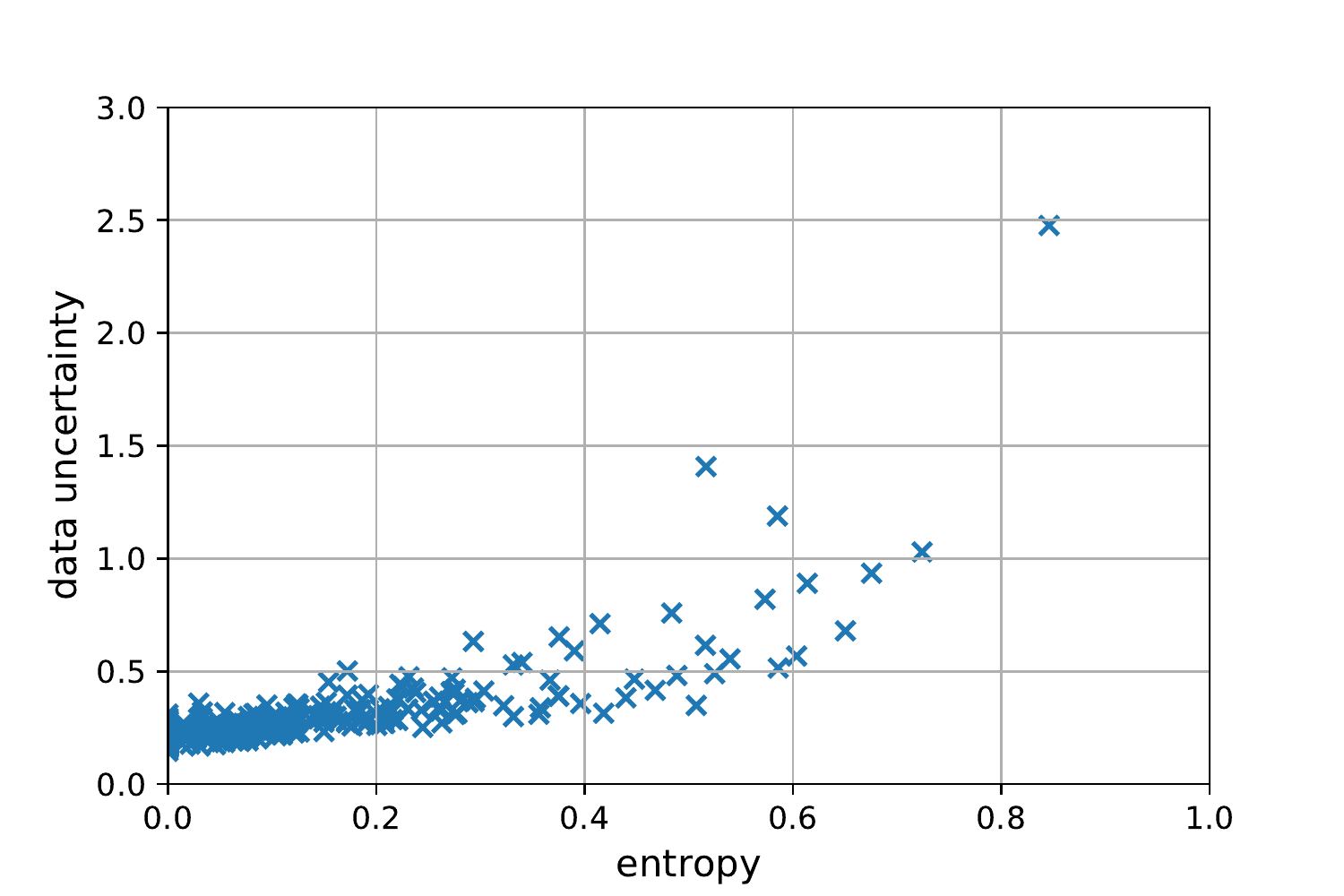}
	\caption{\label{fig:iu_entropy}Scatter plot of evaluated data uncertainty against entropy of annotated NER tag distribution for all tokens in CoNLL 2003 dataset. Higher input-dependent data uncertainties are estimated for input tokens that have higher tag entropies.}
\end{figure}

\subsection{Summary of Results}
We can observe from the results that accounting for uncertainties improves model performances in all three NLP tasks. In detail, for the sentiment analysis setting with CNN models, quantifying both uncertainties gives the best performance and improves upon baseline by up to 26.4\%. For named entity recognition, input-dependent data uncertainty improves F1 scores by 2.7\% in CoNLL 2003. For language modeling, perplexity improves 4.2\% when both uncertainties are quantified.

\section{Analysis}
In the previous section, we empirically show that by modeling uncertainties we could get better performances for various NLP tasks. In this section, we turn to analyze the uncertainties quantified by our approach. We mainly focus on the analysis of data uncertainty. For model uncertainty, we have similar observations to \cite{kendall2017uncertainties}.

\subsection{What Does Data Uncertainty Measure}
In Equation \ref{equ:def_ui}, we define data uncertainty as the proportion of observation noise or variance that is caused by the inputs. Conceptually, input-dependent data uncertainty is high if it is hard to predict its corresponding output given an input. We explore in both sentiment analysis and named entity recognition tasks and analyze the characteristics of inputs with high and low data uncertainties measured by our model.

\begin{figure}\centering    
	\includegraphics[width=3.2in]{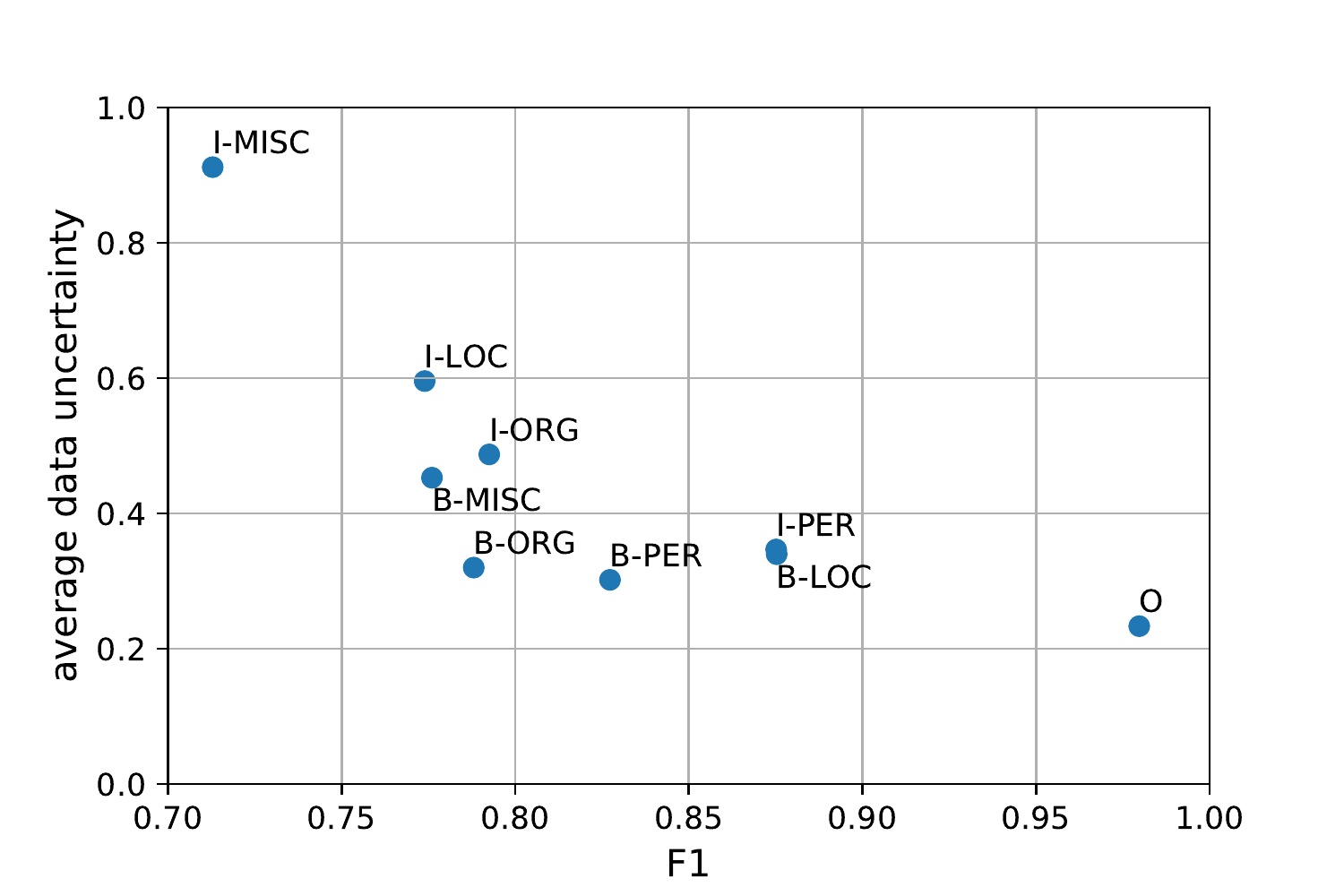}
	\caption{\label{fig:iu_f1}Scatter plot of average evaluated data uncertainty against test set F1 score for different tags. Higher data uncertainties are observed when predicting tags with lower F1 score.}
\end{figure}

Table \ref{tab:sent_iu} shows examples with high and low data uncertainties taken from the Yelp 2013 test set. Due to space limit, we only show four typical examples. Examples with high data uncertainties are either short or very long with extensive descriptions of actions instead of opinions. On the other hand, examples with low data uncertainties are of relatively medium length and contain large amount of strong opinion tokens. These observations are consistent with our intuition.

For the CoNLL 2003 dataset, we take all tokens and measure their average quantified data uncertainty. We use the following strategy to measure how difficult the prediction for each token is: 1. calculate the distribution of NER tags the token is annotated in the training data; 2. use entropy to measure the difficulty level of the prediction defined as:
\begin{align}
H(p_1,p_2,\cdots,p_m)=-\sum_{i=1}^mp_i\log{p_i}
\end{align}
where $p_1, p_2, \cdots, p_m$ is the distribution of NER tags assigned to a particular token in the training set. The higher the entropy, the more tags a token can be assigned and the more even these possibilities are. For example, in the training data, the token \textit{Hong} has been annotated with tag B-LOC (first token in \textit{Hong Kong}), B-ORG, B-PER, B-MISC. Therefore \textit{Hong} has a high entropy with respect to its tag distribution. In contrast, the token \textit{defended} has only been assigned tag O representing outside of any named entities. Therefore \textit{defended} has a low entropy of $0$.

We plot the relationship between the average quantified data uncertainty and NER tag distribution entropy for the tokens in Figure \ref{fig:iu_entropy}. It is clear that for tokens with higher entropy values, data uncertainties measured by our model are indeed higher.

We also analyze the data uncertainty differences among NER tags. For each NER tag, we evaluate its test set F1 score and average data uncertainty quantified by our model. The relationship is shown in Figure \ref{fig:iu_f1}. We observe that when predicting more difficult tags, higher average data uncertainties are estimated by the model. These observations indicate that data uncertainty quantified by our model is highly correlated with prediction confidence.

\section{Conclusion}
In this work, we evaluate the benefits of quantifying uncertainties in modern neural network models applied in the context of three different natural language processing tasks. We conduct experiments on sentiment analysis, named entity recognition, and language modeling tasks with convolutional and recurrent neural network models. We show that by quantifying both uncertainties, model performances are improved across the three tasks.
We further investigate the characteristics of inputs with high and low data uncertainty measures in Yelp 2013 and CoNLL 2003 datasets. For both datasets, our model estimates higher data uncertainties for more difficult predictions.
Future research directions include possible ways to fully utilize the estimated uncertainties. 
\bibliography{aaai}
\bibliographystyle{aaai_bib}
\end{document}